\documentclass[10pt,journal,compsoc]{IEEEtran}

\usepackage[most]{tcolorbox}
\usepackage{epstopdf}
\usepackage{url}
\usepackage{textcomp}
\usepackage{algorithmic}
\usepackage{algorithm}
\usepackage{bm}
\usepackage{amssymb}
\usepackage{multirow}
\usepackage{ragged2e}

\newcommand\rev[1]{{\color{black}{#1}}}

\begin{document}

\title{EdgeAI: A Vision for Deep Learning in IoT Era}

\author{Kartikeya~Bhardwaj,~\IEEEmembership{Member,~IEEE,}
        Naveen~Suda,~\IEEEmembership{Member,~IEEE,}
        and~Radu~Marculescu,~\IEEEmembership{Fellow,~IEEE}
\IEEEcompsocitemizethanks{
\IEEEcompsocthanksitem K. Bhardwaj and N. Suda are with Arm, Inc., San Jose, USA, 95134. Emails: kartikeya.bhardwaj@arm.com, naveen.suda@arm.com
\IEEEcompsocthanksitem R. Marculescu is with Electrical and Computer Engineering Department, Carnegie Mellon University, Pittsburgh, USA, 15213. 
E-mail: radum@cmu.edu
\IEEEcompsocthanksitem (c) 2019 IEEE. Personal use of this material is permitted. Permission from IEEE must be obtained for all other uses, in any current or future media, including reprinting/republishing this material for advertising or promotional purposes, creating new collective works, for resale or redistribution to servers or lists, or reuse of any copyrighted component of this work in other works.
}
}

\markboth{IEEE Design and Test}%
{Bhardwaj \MakeLowercase{\textit{et al.}}: EdgeAI: Challenges and Opportunities}

\IEEEtitleabstractindextext{%
\begin{abstract}
\justify
The significant computational requirements of deep learning present a major bottleneck for its large-scale adoption on hardware-constrained IoT-devices. Here, we envision a new paradigm called \textit{EdgeAI} to address major impediments associated with deploying deep networks at the edge. Specifically, we discuss the existing directions in computation-aware deep learning and describe two new challenges in the IoT era: (1) Data-independent deployment of learning, and (2) Communication-aware distributed inference. We further present new directions from our recent research to alleviate the latter two challenges. Overcoming these challenges is crucial for rapid adoption of learning on IoT-devices in order to truly enable EdgeAI.
\end{abstract}

}

\maketitle
\IEEEdisplaynontitleabstractindextext
\IEEEpeerreviewmaketitle

\IEEEraisesectionheading{\section{Introduction}\label{sec:intro}}
\IEEEPARstart{D}{eep} learning has indeed pushed the frontiers of progress for many computer vision, speech recognition, and natural language processing applications. However, due to their enormous computational complexity, deploying such models on constrained devices has emerged as a critical bottleneck for large-scale adoption of intelligence at the IoT edge. It has been estimated that the number of connected IoT-devices will reach one trillion across various market segments by 2035\footnote{Arm: The Route to a Trillion Devices.\\\url{https://pages.arm.com/route-to-trillion-devices.html}}; this provides us a unique opportunity for integrating widespread intelligence in edge devices. Such an exponential growth in IoT-devices necessitates new breakthroughs in Artificial Intelligence research that can more effectively deploy learning at the edge and, therefore, truly exploit the setup of trillions of IoT-devices. 
Towards this, we envision a new paradigm called \textit{EdgeAI} which aims for widespread deployment of deep learning on IoT: Specifically, to achieve EdgeAI, we must address several major challenges related to the lack of Big Data at the edge, computation-aware deployment of learning models, and communication-aware distributed inference. Overcoming these challenges can enable rapid adoption of intelligence at the edge. 

Perhaps the biggest challenge in the era of edge computing is the constrained hardware of IoT-devices. For instance, IoT-devices typically consist of hundreds of KB of memory and are targeted to run at low operating frequencies for energy efficiency. Since these obvious edge computing challenges like hardware-constraints and security have been surveyed in-depth in~\cite{edgComp}, our focus here will be primarily on challenges for deep learning on edge. To this end, a new area of \textit{model compression} has emerged where the goal is to reduce the size of a pretrained deep network to meet certain energy and latency constraints on edge devices without sacrificing accuracy~\cite{deepComp, hintonKD}. Nevertheless, computational limitations of the edge devices are only a part of the story. In the real-world, deployment of EdgeAI often gets exacerbated by the lack of huge amounts of data at the edge for learning, as well as heavy communication cost among IoT-devices. Hence, we describe below two \textit{new} challenges arising from lack of data on the edge, and communication-aware deep learning.

For many applications such as healthcare via smart wearables, edge devices are deployed in environments that can lack huge amounts of data. We call this the \textit{Learning from Small Data}~\cite{pakdd} which is needed in two scenarios: (\textit{i})~when there is no Big Data for an application and only Small Data exists~\cite{pakdd}, (\textit{ii})~when Big Data is indeed available but is private (\textit{e.g.}, Image Processing on medical images, Speech Recognition, \textit{etc.}). 
Since training on a huge private dataset can result in complex deep learning models, the industries deploying such models at the edge need to compress these deep networks without access to the original training dataset or some alternate unlabeled dataset\footnote{The industry deploying a model at the edge can collect alternate datasets for model compression. However, collecting alternate data may not always be possible, or can be very time consuming/expensive and, hence, infeasible.}. Then, for rapid adoption of deep learning at the edge, an important question is, \textit{how can we perform model compression without using the original, private or alternate datasets?} We refer to this problem as \textit{data-independent model compression}~\cite{ddpaper}. Clearly, data-independent model compression techniques can accelerate the adoption of AI on edge devices because the users trying to deploy a model on IoT-devices will \textit{not} have to rely on the private datasets of third parties. 
\begin{figure*}[!t]
\centering
\includegraphics[width=\textwidth]{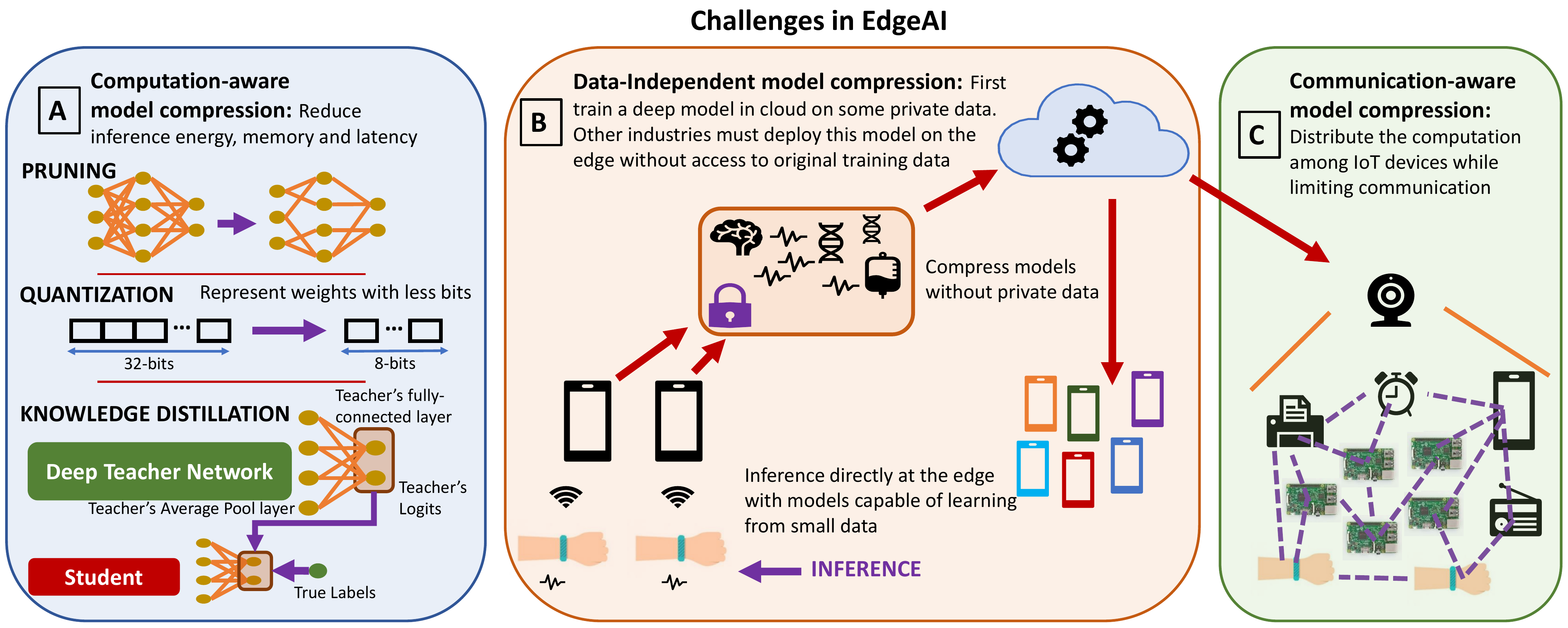}
	\caption{\rev{Major challenges in EdgeAI: (a) Computation-aware model compression: Key techniques include Pruning, Quantization, and Knowledge Distillation (KD)~\cite{deepComp,hintonKD}, (b) Data-independent model compression: Compress deep networks without using private datasets, and (c) Communication-aware model compression: With billions of IoT-devices, their network must be exploited to obtain low-latency, high-accuracy intelligence at the edge.}}
\label{flow}
\end{figure*}

Finally, since IoT naturally implies a network of connected devices, it automatically opens the door to a new class of problems -- \textit{communication-aware model compression}~\cite{nonnpaper}. For instance, many smart home/cities applications can have numerous connected IoT-sensors with, say, 500KB total memory per node. As the compressed deep networks often grow in size to achieve higher accuracy, due to strict memory-constraints, such models must be distributed across multiple devices; this generates significant inter-device communication at each layer of the deep network. Hence, a massive communication cost arising from this distributed inference presents another major (and largely ignored) impediment for widespread deployment of deep learning on IoT. Therefore, new ways must be found to compress models for IoT-devices which reduces both computation and communication costs. In other words, since the IoT paradigm consists of connected devices, this massive \textit{network} of billions of devices must be exploited to obtain highly accurate and low-latency intelligence at the edge.

In view of the above, we identify three major challenges faced by deployment of EdgeAI: (\textit{i})~Computation-aware learning on IoT \rev{(see Fig.~\ref{flow}(a))}, (\textit{ii})~\rev{Data-independent model compression for Learning from Small Data (see Fig.~\ref{flow}(b))}, and (\textit{iii})~Communication-aware deployment of deep learning models on multiple IoT devices \rev{(see Fig.~\ref{flow}(c)). For all these challenges, we review the current work and summarize our latest research ideas. The objective of this \textit{vision} paper is to (1) particularly highlight the importance of the latter two, \textit{new} challenges for rapid adoption of AI at the edge, and (2) unify them under the EdgeAI paradigm. We believe, these challenges will lead to future research directions in this field.}

%
\section{Computation-Aware Model Compression}
Power- and memory-constrained hardware of IoT-devices continues to be the biggest challenge for EdgeAI. Towards this, the field of model compression has received a lot of attention to make deep learning models more suitable for edge devices~\cite{deepComp, hintonKD}. Besides model compression, another venue of active research is hardware accelerators and codesign of models and hardware architectures for efficient deep learning inference. Since our focus is on challenges for Artificial Intelligence for edge computing, we will mostly focus on model compression in this article. 

Model compression refers to a class of techniques that reduce the size (\textit{i.e.}, number of parameters) and computations (\textit{i.e.}, Floating Point Operations, FLOPS) of deep networks without losing accuracy. Most of the prior art in model compression literature focuses only on the computational aspects such as power, latency, memory, and energy consumption. Specifically, as summarized in Fig.~\ref{flow}(a),
there are three main directions for model compression:
\begin{itemize}
\item \textbf{Pruning:} These refer to model compression methods where the redundant or useless weights or even channels are completely removed~\cite{deepComp}. Pruning does \textit{not} reduce the number of layers in the original model.
\item \textbf{Quantization:} Conventionally, the deep networks are trained with 32-bit floating point weights and activations. Quantization techniques reduce the number of bits used for representing weights and activations, thereby reducing the memory footprint of models~\cite{deepComp}.
\item \textbf{Knowledge Distillation (KD):} KD trains a significantly smaller student network to mimic a large teacher model (which we want to compress). KD allows us to directly reduce the number of layers compared to the teacher model~\cite{hintonKD}. KD has also been shown to work with unlabeled datasets.
\end{itemize}

In the rest of the paper, we address two new challenges to show that computation-aware model compression alone cannot enable the true potential of EdgeAI.

\section{\rev{Data-Independent Model Compression}}
When Big Data is indeed available but is private, the industries trying to deploy such models on the edge cannot use the original datasets for model compression. 
We discuss this important case below.

\subsection*{Dream Distillation}
Consider a KD-based model compression problem where a teacher model is trained on CIFAR-10 dataset (see Fig.~\ref{flow}(a)). It has been shown that \textit{alternate datasets} such as CIFAR-100 can be used to train a student model via KD~\cite{mismatch}. Specifically, during distillation, the teacher can transfer relevant knowledge about CIFAR-10 to the student even when trained with alternate datasets like CIFAR-100. The resulting student can demonstrate reasonably good accuracy on the original intended task, \textit{i.e.}, the CIFAR-10 classification. 
Since in many cases, it may not be possible to obtain even the alternate real datasets (see Footnote~2), an important question is \textit{how can we compress a deep network without using the original training set or any alternate real data, while achieving comparable accuracy?} \rev{To answer this question, we present our recent work on \textit{Dream Distillation}~\cite{ddpaper}}, a KD-based model compression technique which does not rely on access to \textit{any} real data.

\subsubsection*{Assumptions}
\begin{figure}[!t]
\centering
\includegraphics[width=3.4in]{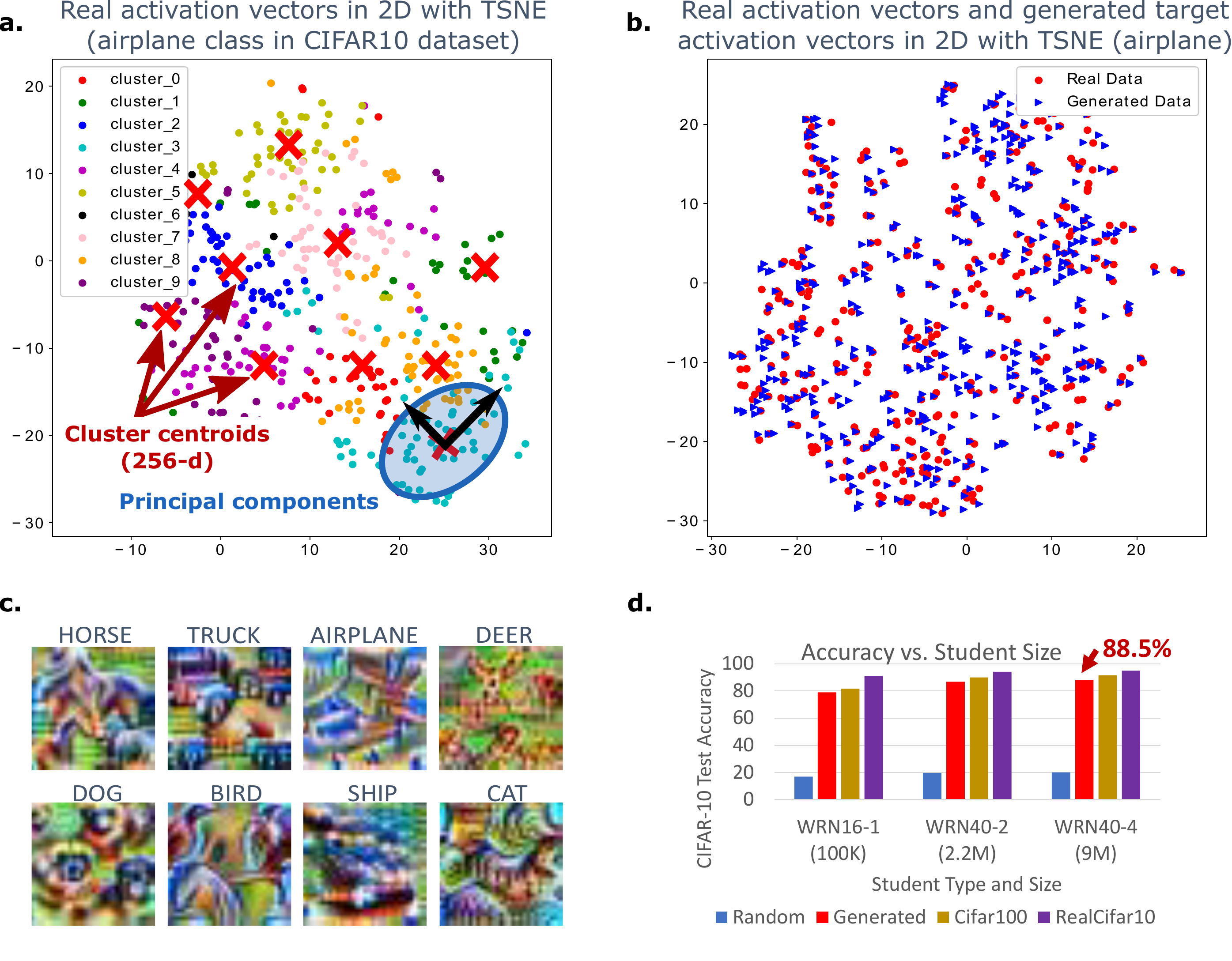}
	\caption{Dream Distillation: A data-independent knowledge distillation framework~\cite{ddpaper}. (a) Metadata used by our method, and (b) tSNE visualization of real data activations and target activations generated, (c) Examples of generated samples for CIFAR-10 dataset, and (d) Accuracy of student models trained from random data, generated data, alternate real data: CIFAR100, and real CIFAR-10 dataset.}
\label{dd}
\end{figure}

We assume a large Wide Resnet (WRN40-4, $8.9$M parameters) as the teacher network, and a small Wide Resnet (WRN16-1, 100K parameters) as the student model. When trained, the teacher model achieves about $95\%$ accuracy on CIFAR-10 dataset. Also, when the student WRN16-1 model is trained via Attention Transfer-based KD (ATKD)~\cite{atkd} with WRN40-4 teacher, the student model achieves about $91\%$ accuracy while using $89\times$ fewer parameters.  We further assume that no real data is available (\textit{i.e.}, neither alternate data, nor original training set) but only some small amount of metadata is available, as described below:
\begin{enumerate}
\item We are given \textit{cluster-centroids} created from $k$-means clustering of real activation vectors at the average-pool layer\footnote{Average-pool layer refers to the averaged output of final convolutional layer of a deep network (see Fig.~\ref{flow}(a)).} of the teacher network for $10\%$ of the real CIFAR-10 images. Fig.~\ref{dd}(a) illustrates the cluster-centroids for ten clusters present in the airplane class for CIFAR-10.
\item Principal components for each cluster are also available as part of our metadata (see Fig.~\ref{dd}(a)).
\end{enumerate}

Therefore, our metadata does not contain any identifying information about the real images (as only mean activations are used). 
The core idea of our approach is to use this metadata and the pretrained teacher network to generate a new dataset of synthetic images which can effectively distill relevant knowledge from the teacher to the student without using real images. 

\rev{A relevant prior work is Data-Free Knowledge Distillation (DFKD)~\cite{kd123} which also relies on some metadata. However, DFKD specifically states that using metadata from only one layer of the teacher model leaves the problem under-constrained and, hence, leads to very poor accuracy (\textit{e.g.}, $68$-$77\%$ accuracy on MNIST dataset). Consequently, DFKD will achieve significantly lower accuracy for much more complex datasets like CIFAR-10. In contrast, we precisely demonstrate that metadata from a single layer is sufficient to achieve high accuracy~\cite{ddpaper}.}

\subsubsection*{Dream Generation and Distillation}
Our approach is rooted in the well-known field of deep learning interpretability where the goal is to understand and visualize what different neurons in deep networks are learning\footnote{Chris Olah \textit{et al.} The building blocks of interpretability. Webpage:\\ \url{https://distill.pub/2018/building-blocks/}.}. The field of visualization in deep learning was popularized by the initial work on DeepDream\footnote{DeepDream: \url{https://bit.ly/2Nd9eae}} and feature visualization tools such as Tensorflow Lucid\footnote{Tensorflow Lucid: \url{https://github.com/tensorflow/lucid}}. Feature visualization generates an image that can maximize a certain objective. For instance, we can use the above tools to generate an image that can maximally activate a given hidden unit (\textit{e.g.}, a neuron or a channel) at a certain hidden layer. 
In DeepDream, these generated images are called the \textit{Dreams} of the deep network. This is where the term Dream Distillation comes from, as if we are distilling what the teacher dreams!

For Dream Distillation, we create custom objectives from the metadata. Specifically, we first generate target activations from the metadata by adding a small amount of noise to the cluster-centroids along the directions of principal components. As demonstrated in Fig.~\ref{dd}(b), a two-dimensional visualization\footnote{Two-Dimensional Visualization via tSNE: \url{https://bit.ly/2FUmCzj}} of target activations (blue triangles) and real data activations (red circles) at the average-pool layer of the teacher network shows that their distributions are quite similar. Next, we generate $50,000$ images by using Tensorflow Lucid with the objective that activations of generated images at teacher's average-pool layer must be as close as possible to these target activations. Finally, these generated (synthetic) images are used for distilling the knowledge from teacher to the student. \rev{More details are given in~\cite{ddpaper}.}

Fig.~\ref{dd}(c) shows some of the samples generated by our \textit{Dream Generation} technique described above. Clearly, while for classes like \{horse, truck, deer, dog\}, some key features (\textit{e.g.}, distorted-animal-faces, wheels, \textit{etc.}) of these objects are revealed, for other classes such as \{ship, airplane, cat, bird\}, the images are much more subtle. For example, for cat samples, the teacher network mostly generates a striped pattern which is typical of many cats, but may not be the most identifying feature of a cat from a human point-of-view. Then, the question is, \textit{can these images be used for distilling knowledge from teacher to the student?}

To answer the above question, we use four different datasets containing $50,000$ images for experiments in Fig.~\ref{dd}(d): (\textit{i}) Random noise images (blue), (\textit{ii}) Dream Distillation images (red), (\textit{iii}) CIFAR-100 images as an alternate real dataset (yellow), and (\textit{iv}) Real CIFAR-10 dataset (violet). We use three student models of different sizes: WRN16-1 (100K parameters), WRN40-2 (2.2M parameters), and WRN40-4 (8.9M parameters). 
As evident from Fig.~\ref{dd}(d), Dream Distillation performs on par with the alternate real dataset CIFAR-100, both of which are reasonably close to the real CIFAR-10 student accuracy. 
More specifically, for WRN16-1 student, generated images achieve about $79\%$ accuracy, while CIFAR-100 images achieve around $81\%$ accuracy on CIFAR-10 test set. Moreover, we further demonstrate that the WRN40-4 student trained via Dream Distillation achieves $88.5\%$ accuracy on CIFAR-10 test set without ever seeing \textit{any} real data!
%
This shows that the synthetic images generated via our method can be used to transfer significant amount of relevant knowledge about the real data without accessing any real or alternate datasets. Therefore, Dream Distillation can have great implications as it can allow industries to more rapidly deploy EdgeAI without access to third-party proprietary data.






\section{Communication-Aware Model\\Compression}
Most of the prior work in model compression focuses on computational aspects such as energy, memory, and latency of inference on a \textit{single} device~\cite{deepComp, hintonKD}. In contrast, the literature that addresses communication-aware deployment of deep learning models on \textit{multiple} devices is significantly more limited. 
Recent models for distributed inference such as SplitNet~\cite{splitnet} and MoDNN~\cite{modnn} do not use ideas from model compression (\textit{e.g.}, pruning/distillation). For instance, SplitNet aims to split a deep learning model into disjoint subsets during training. However, it does \textit{not} consider the memory- and FLOP-budgets for IoT-devices~\cite{splitnet}. Consequently, the disjoint parts of the network obtained by SplitNet do not satisfy the strict memory-constraints of individual devices. Similarly, MoDNN~\cite{modnn} reduces the number of FLOPS during distributed inference. However, MoDNN assumes that the entire model can fit on each device, and tries to deploy a model directly on the edge without model compression. 

Since many IoT-devices are significantly memory-constrained (\textit{e.g.}, microcontrollers with 500KB memory), the assumption that the entire model can fit on each IoT-device is very optimistic. Indeed, in such a memory-constrained scenario, the model itself must be distributed onto multiple devices which will lead to heavy communication among devices at each layer of the deep network. Therefore, we need a new model compression paradigm that not only reduces memory and computation of deep networks but also minimizes communication among multiple devices for efficient distributed inference. With ever-more devices available on the edge, this network of devices must be exploited to improve model accuracy without increasing the communication latency. \rev{Towards this, we describe our recent work on Network-of-Neural Networks (NoNN)~\cite{nonnpaper}} for memory- and communication-aware model compression.

\subsection*{Network-of-Neural Networks}
We follow a KD-based teacher-student model compression approach (see Fig.~\ref{flow}(a)). A NoNN refers to a new framework which allows for memory- and communication-aware student architectures obtained from a single powerful teacher model. Essentially, a NoNN represents a collection of multiple student modules which focus only on a part of teacher's knowledge. Since training an individual student to mimic a specific partition of teacher's knowledge is significantly easier than mimicking teacher's entire knowledge, NoNN achieves higher accuracy with very limited communication among the students. Distributed inference can then be performed by deploying individual student modules on different resource-constrained devices. 

\begin{figure}[]
\centering
\includegraphics[width=3.4in]{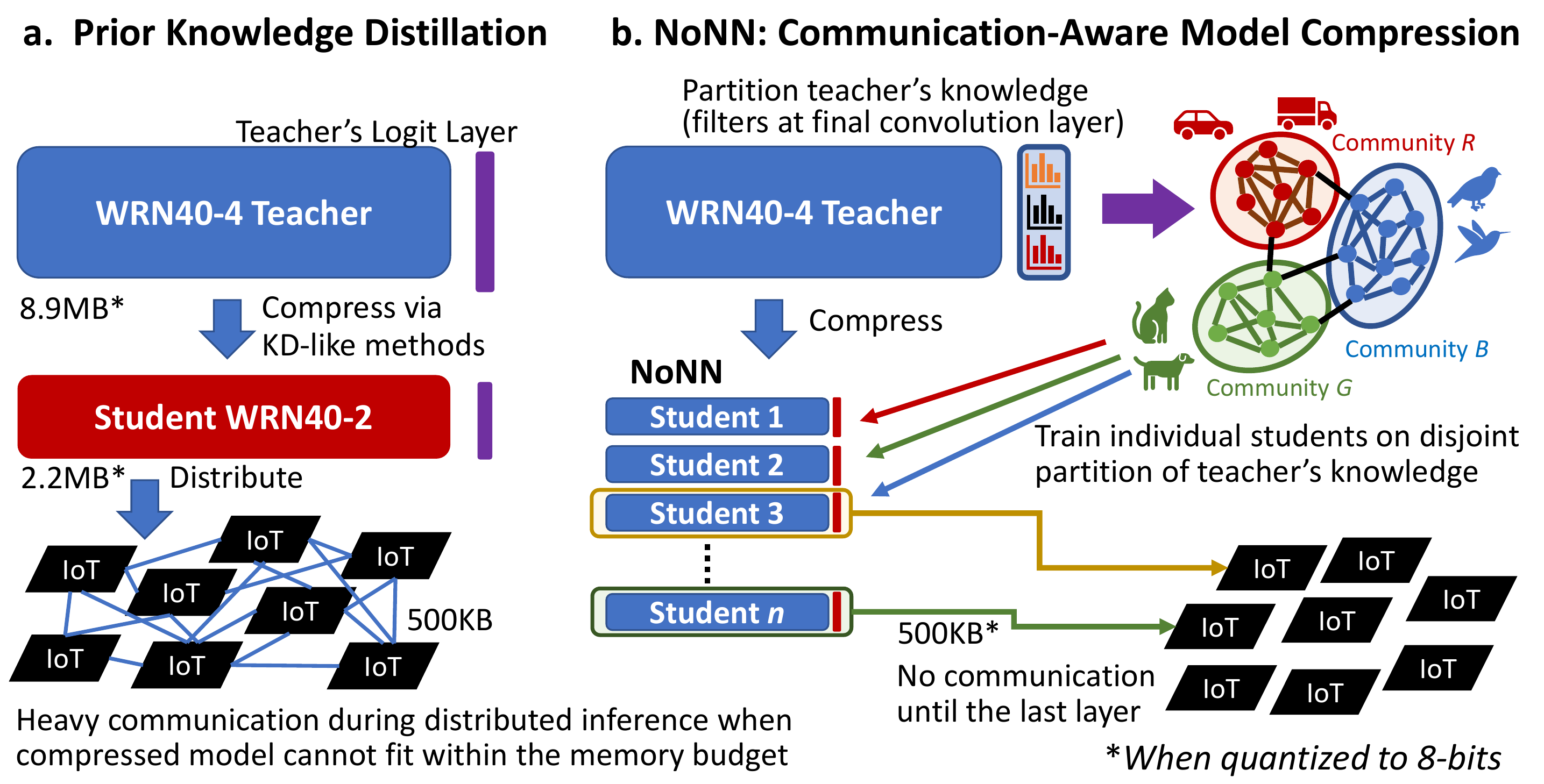}
	\caption{(a) Prior KD~\cite{hintonKD}: Distributing large student models that do not fit on a single memory-limited IoT device leads to heavy communication at each layer. (b) NoNN~\cite{nonnpaper} results in disjoint students that can fit on individual devices: \rev{Filters at teacher's final convolution layer (representing knowledge about different classes) can be partitioned to train individual students which results in minimal communication until the final layer.}}
\label{nonn}
\end{figure}

Fig.~\ref{nonn} describes the difference between traditional KD and NoNN. As shown, NoNN uses \textit{network science}~\cite{networksci} to partition teacher's final convolution layer \rev{since each student is trained on a disjoint part of teacher's knowledge. Specifically, in a deep Covolutional Neural Network (CNN), different filters activate for different classes. These \textit{patterns of activations} reveal how knowledge learned by the teacher network is distributed at the final convolution layer. Therefore, we first use these patterns to create a \textit{filter activation network}~\cite{nonnpaper} which represents how knowledge is distributed across multiple filters (see Fig.~\ref{nonn}(b)). We then partition this network into disjoint subsets via community detection~\cite{networksci}, a network partitioning technique (see~\cite{nonnpaper} for details).}

Once teacher's knowledge is partitioned, we train completely separate students to mimic only a part of teacher's function. This results in a highly parallel student architecture as shown in Fig.~\ref{nonn}(b). Similar to KD, since we can select significantly smaller individual student modules (subject to some memory/FLOP budgets), our NoNN student model results in significantly lower memory, computations, as well as communication. In fact, the individual student models do not communicate with each other until the final fully connected layer!

As a proof-of-concept, we experiment with NoNN for CIFAR-10 dataset. We assume that each IoT-device has a memory budget of 500KB, \textit{i.e.}, each individual student module in Fig.~\ref{nonn}(b) must have less than 500K parameters (since a model with 500K parameters can fit within 500KB memory when quantized to 8-bits). Our teacher model for this experiment is \rev{again Wide Resnet WRN40-4 (8.9M parameters), and we focus on a NoNN with two student modules, each based on a Wide Resnet structure (denoted NoNN-2S)~\cite{nonnpaper}}. Overall, our NoNN-2S model has about 860K parameters, while each student module has about 430K parameters. \rev{Using TVM\footnote{\rev{TVM compiler: \url{https://tvm.ai/}}}}, we further deploy all models on Raspberry-Pi (RPi) devices which nicely represent an edge scenario. We put individual students on separate RPi's (communicating via a point-to-point wired connection) to demonstrate the effectiveness of NoNN. 

Table~\ref{nt}(top) shows the results for the teacher model deployed on one RPi and our NoNN-2S model deployed on two RPi's. As evident, we very significantly outperform the teacher model ($13\times$ to $20\times$ improvement) in memory, FLOPS, and energy with only $1\%$ loss of accuracy. Note that, our theoretical improvement (per student) in FLOPS is about $15\times$ and the practical improvement in energy is $14\times$; this shows great agreement between theory and practice. \rev{In~\cite{nonnpaper}, we conduct extensive experiments to demonstrate the effectiveness of NoNN on five well-known image classification tasks. We also show results for a higher number of students.} This clearly shows the superiority of NoNN.
\begin{table}[]
\rev{
\centering
\caption{\rev{Communication-Aware Model Compression: CIFAR-10 Results~\cite{nonnpaper}}}
\scalebox{0.897}{
\label{nt}
\begin{tabular}{|l|c|c|c|} 
\hline
\multirow{2}{*}{TVM Setup}& WRN40-4 Teacher & NoNN-2S (one student & \multirow{2}{*}{Gain}\\
& (on a single RPi)~ & on each RPi)$^*$& \\
\hline \hline
$\#$parameters& $8.9$M& $\bm{0.43}$\textbf{M}& $\bm{20.7\times}$\\ \hline
$\#$FLOPS& $2.6$G& $\bm{167}$\textbf{M}& $\bm{15.5\times}$\\ \hline
Latency (ms)& $1405$& $\bm{115}$& $\bm{12.2\times}$\\ \hline
Energy (mJ)& $3430.67$& $\bm{238.98}$& $\bm{14.3\times}$\\ \hline
Accuracy & $\bm{95.49\%}$& $94.32\%$& $-1.17\%$\\ \hline \hline
\end{tabular}
}
\scalebox{0.896}{
\begin{tabular}{|l|c|c||c|}
\multirow{2}{*}{Pytorch Setup} & \multicolumn{2}{c||}{~Split-ATKD (WRN40-2)} & ~NoNN-8S~ \\ \cline{2-3} 
 & 4 RPi's & 8 RPi's & 8 RPi's\\ \hline \hline
Accuracy & $95.03\%$ & $95.03\%$ & $95.02\%$ \\ \hline
Parameters per device & $550$K & $275$K & $430$K \\ \hline
FLOPS per device & $163$M & $82$M & $167$M \\ \hline
Total latency per inference (s)~ & $23$ & $28.5$ & $\bm{0.85}$ \\ \hline \hline
Speedup with NoNN & $\bm{27\times}$ & $\bm{33\times}$ & $-$ \\ \hline
\end{tabular}
}
}
\begin{flushleft}
{\scriptsize $^{*}$These results are for each individual student module. For complete NoNN, $\#$parameters will be more than $0.8$M, and FLOPS and total energy (for both RPi's) will double. However, this is acceptable since we are mostly concerned with per-device budgets for memory, FLOPS, and energy.\vspace{-3mm}}
\end{flushleft}
\end{table}

Finally, we verify that a deep network split horizontally will lead to heavy communication at each layer. 
\rev{Towards this, we train a NoNN-8S model for deployment on eight RPi's. For comparison, we distribute an ATKD-based compressed model (WRN40-2, 2.2M parameters, $95.03\%$ accuracy)~\cite{atkd} on four and eight RPi's. Note that, we conduct all splitting experiments in Pytorch (including for NoNN-8S) since TVM binary cannot be distributed across multiple devices~\cite{nonnpaper}. As shown in Table~\ref{nt}(bottom), even though FLOPS per device executed for split-ATKD are significantly lower, NoNN-8S is $27$-$33\times$ faster than the split-ATKD models, and achieves a similar accuracy of $95.02\%$. Hence, when the compressed models cannot fit within the given memory-budget}, communication can indeed hamper the inference latency significantly. This clearly highlights the importance of our communication-aware model compression.

\section{Conclusion}
In this paper, we have introduced a new paradigm called \textit{EdgeAI} that aims to extensively deploy deep learning models on the edge. Towards this, we have first discussed existing challenges in computation-aware model compression. Then, we have presented two new challenges related to lack of data for learning on the edge (\textit{i.e.}, data-independent model compression), and communication-aware model compression.  To meet the latter two challenges, we have also presented new directions from our recent research: Dream Distillation for data-independent model compression, and Network-of-Neural Networks for communication-aware model compression. Overcoming all of the above challenges can truly enable widespread adoption of EdgeAI.

%



\ifCLASSOPTIONcaptionsoff
  \newpage
\fi



%

\bibliographystyle{abbrv}
\bibliography{sample-bibliography}

\begin{thebibliography}{10}

\bibitem{nonnpaper}
K.~Bhardwaj, C.~Lin, A.~Sartor, and R.~Marculescu.
\newblock {Memory- and Communication-Aware Model Compression for Distributed
  Deep Learning Inference on IoT}.
\newblock In {\em Proceedings of the International Conference on
  Hardware/Software Codesign and System Synthesis (CODES+ISSS), ESWEEK-TECS
  special issue, \textit{To Appear}, arXiv preprint arXiv:1907.11804}, pages
  1--22, Oct. 2019.

\bibitem{pakdd}
K.~Bhardwaj and R.~Marculescu.
\newblock Dimensionality reduction via community detection in small sample
  datasets.
\newblock In {\em Pacific-Asia Conference on Knowledge Discovery and Data
  Mining (PAKDD)}, pages 102--114. Springer, 2018.

\bibitem{ddpaper}
K.~Bhardwaj, N.~Suda, and R.~Marculescu.
\newblock {Dream Distillation: A Data-Independent Model Compression Framework}.
\newblock {\em ICML 2019 Joint Workshop on On-Device Machine Learning and
  Compact Deep Neural Network Representations (ODML-CDNNR), \textbf{Oral
  Presentation}, arXiv preprint arXiv:1905.07072}, pages 1--4, Jun. 2019.

\bibitem{deepComp}
S.~Han, H.~Mao, and W.~J. Dally.
\newblock Deep compression: Compressing deep neural networks with pruning,
  trained quantization and huffman coding.
\newblock {\em arXiv preprint arXiv:1510.00149}, 2015.

\bibitem{hintonKD}
G.~Hinton, O.~Vinyals, and J.~Dean.
\newblock Distilling the knowledge in a neural network.
\newblock {\em arXiv preprint arXiv:1503.02531}, 2015.

\bibitem{splitnet}
J.~Kim, Y.~Park, G.~Kim, and S.~J. Hwang.
\newblock Splitnet: Learning to semantically split deep networks for parameter
  reduction and model parallelization.
\newblock In {\em International Conference on Machine Learning}, pages
  1866--1874, 2017.

\bibitem{mismatch}
M.~Kulkarni, K.~Patil, and S.~Karande.
\newblock Knowledge distillation using unlabeled mismatched images.
\newblock {\em arXiv preprint arXiv:1703.07131}, 2017.

\bibitem{kd123}
R.~Lopes, S.~Fenu, and T.~Starner.
\newblock Data-free knowledge distillation for deep neural networks.
\newblock {\em arXiv preprint arXiv:1710.07535}, 2017.

\bibitem{modnn}
J.~Mao and et~al.
\newblock Modnn: Local distributed mobile computing system for deep neural
  network.
\newblock In {\em 2017 DATE Conference}, pages 1396--1401. IEEE, 2017.

\bibitem{networksci}
M.~Newman, A.-L. Barabasi, and D.~J. Watts.
\newblock {\em The structure and dynamics of networks}, volume~19.
\newblock Princeton University Press, 2011.

\bibitem{edgComp}
W.~Shi, J.~Cao, Q.~Zhang, Y.~Li, and L.~Xu.
\newblock Edge computing: Vision and challenges.
\newblock {\em IEEE Internet of Things Journal}, 3(5):637--646, 2016.

\bibitem{atkd}
S.~Zagoruyko and N.~Komodakis.
\newblock Improving the performance of convolutional neural networks via
  attention transfer.
\newblock {\em ICLR}, 2017.

\end{thebibliography}

%
%

%


%
%




\end{document}